# Distributional Estimation of Data Uncertainty for Surveillance Face Anti-spoofing

Mouxiao Huang[1, 2, *]

[1] The Guangdong Provincial Key Laboratory of Computer Vision and Virtual Reality Technology, Shenzhen Institute of Advanced Technology, Chinese Academy of Sciences

[2] University of Chinese Academy of Sciences

Shenzhen, China

[*] mx.huang@siat.ac.cn

*Abstract*—Face recognition systems have become increasingly vulnerable to security threats in recent years, prompting the use of Face Anti-spoofing (FAS) to protect against various types of attacks, such as phone unlocking, face payment, and self-service security inspection. While FAS has demonstrated its effectiveness in traditional settings, securing it in long-distance surveillance scenarios presents a significant challenge. These scenarios often feature low-quality face images, necessitating the modeling of data uncertainty to improve stability under extreme conditions. To address this issue, this work proposes Distributional Estimation (DisE), a method that converts traditional FAS point estimation to distributional estimation by modeling data uncertainty during training, including feature (mean) and uncertainty (variance). By adjusting the learning strength of clean and noisy samples for stability and accuracy, the learned uncertainty enhances DisE's performance. The method is evaluated on SuHiFiMask [1], a large-scale and challenging FAS dataset in surveillance scenarios. Results demonstrate that DisE achieves comparable performance on both ACER and AUC metrics.

## I. Introduction

As face recognition technology continues to become more prevalent in security and surveillance systems [2], the risk of face spoofing attacks poses a significant threat to the effectiveness of these systems [8]. Face spoofing attacks involve using fake faces to trick the system into granting access or authentication. These attacks come in various forms such as replay-attack, print-attack, and face-mask. Replay-attack involves the use of previously recorded videos or images of an authorized person, while print-attack utilizes printed copies of a face to bypass the system. In recent times, the use of face masks to carry out spoofing attacks has become a significant concern, especially in the wake of the COVID-19 pandemic. Consequently, developing reliable Face Anti-Spoofing (FAS) methods that can effectively detect and prevent such attacks has become a crucial area of research, requiring constant innovation and improvement to stay ahead of ever-evolving threats.

Initially, FAS methods relied on handcrafted features [3], which required prior knowledge and human liveness cues to differentiate between live and spoof faces. Although significant progress has been made in the performance of face presentation attack detection technology in short-distance applications [4] such as face payment, and self-service security inspection, it still struggles with face quality and cannot be applied to long-distance scenarios. This shortcoming obstructs the deployment of FAS in surveillance scenario. With the advent of deep learning, FAS methods based on deep neural networks have achieved state-of-the-art performance. Nevertheless, most existing methods remain susceptible to unknown spoofing attacks. As the use of surveillance systems in long-distance scenarios continues to increase, detecting face spoofing attacks has become an even more significant challenge [1]. In these scenarios, low-quality faces are common, and they often lack the details needed for fine-grained feature-based FAS tasks. Therefore, the development of effective FAS systems capable of handling these challenging scenarios is critical to ensure the security and reliability of surveillance applications.

FAS methods have become increasingly important due to the rising concern of face spoofing attacks in various security and surveillance systems [2] that incorporate face recognition technology [8]. These attacks can range from simple replay-attacks to more sophisticated methods such as print-attacks and face-mask spoofing. Traditional FAS methods have typically relied on deterministic models that provide a point embedding without considering data uncertainty, resulting in a lack of stability and reliability. However, recent research in face recognition has recognized the importance of modeling data uncertainty to improve the accuracy and robustness of FAS methods. Various methods have been developed to model data uncertainty, such as PFE [6], which estimates a Gaussian distribution in the latent space to make the uncertainty learnable. DUL [7] extends this approach to make both the feature and uncertainty learnable. RTS [8], on the other hand, models the stochastic distribution of temperature, which depends on the input data, and can learn both the face representation and temperature simultaneously, acting like a Bayesian network [12]. These methods have demonstrated improved performance in face anti-spoofing, highlighting the significance of modeling data uncertainty.

Our proposed method, Distributional Estimation (DisE), addresses the issue of low-quality face images by converting traditional point estimation to distributional estimation through modeling data uncertainty. DisE models data uncertainty with a Gaussian distribution in the latent space, and the uncertainty, learned during training, adjusts the learning strength of clean and noisy samples by acting as temperature scaling in the Softmax function. Experimental results show the effectiveness of our method in detecting various types of face spoofing attacks, particularly in extreme conditions with low-quality images. The main contributions of this paper are summarized below:

- The proposed Distributional Estimation (DisE) method converts traditional point estimation to distributional estimation by modeling data uncertainty, which improves the stability and reliability of face anti-spoofing (FAS) methods, especially in extreme conditions with low-quality face images.
- DisE models data uncertainty using a Gaussian distribution in latent space, and the learned uncertainty plays a role in temperature scaling in the Softmax function, adjusting the learning strength of both clean and noisy samples during training.
- The experimental results demonstrate the effectiveness of the proposed DisE method in detecting various types of face spoofing attacks, providing a significant contribution to the development of FAS methods.

## II. RELATED WORK

### A. Face Anti-spoofing Methods

Various methods have been proposed for Face Anti-spoofing to ensure reliable and secure facial recognition systems. These include traditional handcrafted feature-based methods and deep learning-based methods, with promising results shown by CNN-based methods. Traditional methods were designed based on human liveness cues, such as gaze tracking, facial or head movements and eye-blinking, for dynamic discrimination. However, challenges remain, particularly in adapting to different scenarios and attack types. To address these challenges, there is a need to develop more robust and efficient FAS methods. In surveillance scenarios, several face recognition datasets have been released to improve the representation of the global population. Face recognition algorithms utilizing adversarial generative networks and fully convolutional architectures have been developed for surveillance scenarios. Additionally, SFace [10] proposed a sigmoid-constrained hypersphere loss, while AdaFace [11] introduced an adaptive marginal function to prioritize the role of clean samples in classification.

### B. Data Uncertainty in Face Recognition

Deep learning systems often face uncertainty, which can be classified into two categories: data and model uncertainty. Data uncertainty reflects the quality of input data, and recent studies have incorporated it into face recognition. These techniques have investigated how uncertainty influences in deep learning systems. For instance, PFE [6] estimates a Gaussian distribution in the latent space, allowing for learnable uncertainty. PFE's primary drawback is its dependence on pre-trained face representations. DUL [7] extends this approach to enable both the feature and uncertainty to be learnable. While DUL adds uncertainty score to the original model directly. The introduction of randomness in the representation of DUL creates ambiguity in the learning objective. Furthermore, sampling from the high-dimensional distribution is not an efficient strategy for training. RTS [8] models the stochastic distribution of temperature, which depends on input data, and learns the face representation and temperature simultaneously, resembling a Bayesian network. While these approaches have only focused on recognition modules, incorporating data uncertainty into FAS can enhance its robustness.

## III. METHODOLOGY

### A. Problem Formulation

The task of Face Anti-spoofing (FAS) is framed as a binary classification problem where an input face image is categorized as belonging to either an authorized or unauthorized individual. It can be defined mathematically as follows: Given a face image $x$, FAS aims to learn a function $F = f(x)$ that maps the input image to a binary output $y \in \{0,1\}$, where $y = 0$ denotes a genuine face and $y = 1$ denotes a spoofed face. The FAS model can be trained by minimizing a suitable loss function $L(f(x), y)$ over a training dataset $D$ of face images and their corresponding labels, where $L$ is a binary classification loss function, such as cross-entropy (CE) loss. The goal is to learn a model $F$ that can generalize well to unseen face images and accurately classify them as genuine or spoofed.

### B. Distributional Estimation of Data Uncertainty

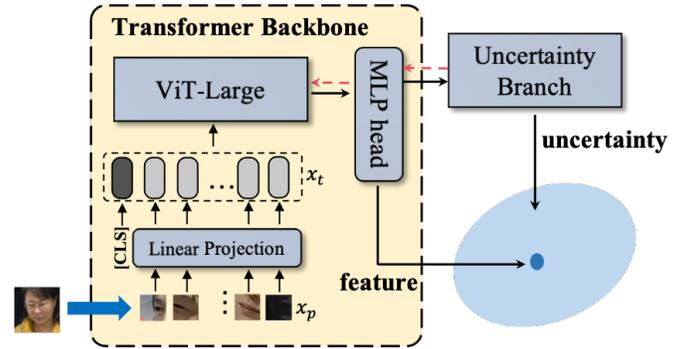

Figure 1. Overview of the proposed Distribution Estimation (DisE). Our method trains the feature and uncertainty simultaneously, and the uncertainty scores adjust the learning strength of different quality samples.

In surveillance scenarios where low-quality faces are prevalent, it is challenging to perform fine-grained feature-based face anti-spoofing (FAS) tasks due to the lack of details in the images. To address this issue, we propose a method called Distributional Estimation (DisE) that introduces data uncertainty quantification into FAS. DisE converts the original point estimation task to distributional estimation by modeling the latent space as a Gaussian distribution, where the mean and variance correspond to the feature and uncertainty, respectively. Figure 1 illustrates that DisE trains the feature and uncertainty concurrently, and the uncertainty scores adjust the learning strength of different quality samples. Specifically, the transformer backbone predicts the feature embedding of the input image, and an extra uncertainty branch predicts the uncertainty score. This branch has a linear layer and a batch normalization layer BN with weights $w_G$ and bias $b_G$, and it follows an exponential function and a Softplus function. The uncertainty branch G predicts the uncertainty, which is equivalent to the variance of the Gaussian distribution in the latent space. Mathematically, the branch can be formulated as:

$$G = \ln(1 + \exp(\exp(BN(w_G * x + b_G)))) \quad (1)$$

Our proposed method employs a modified cross-entropy (CE) loss function to train the classification transformer backbone network. The CE loss can be defined as below:

$$CE = -\log\left(\frac{e^{\frac{s}{v}}}{\sum e^{\frac{s}{v}}}\right) \quad (2)$$

where $v$ refers to the learnable variance, which represents the uncertainty score, and $s$ is the predicted feature. In addition to learning the feature embedding, we apply KL divergence regularization, following the RTS approach, to constrain the uncertainty score and ensure that the distribution $N(s,v)$ conforms to a Gaussian distribution $N(0,I)$ in latent space by minimizing the distance of these two distributions.

## IV. EXPERIMENTS

### A. Experimental Setup

#### 1) Datasets

To evaluate the effectiveness of our proposed approach in surveillance settings, we employed a large-scale dataset called SuHiFiMask [1] as our primary dataset. SuHiFiMask consists of 40 real-life surveillance scenes, such as movie theaters, security gates, and parking lots, which represent a diverse range of face recognition scenarios. It includes 101 participants of different ages and genders, performing various natural activities of daily life. Additionally, the dataset contains several types of spoofing attacks, such as high-fidelity masks, 2D attacks, and adversarial attacks. The data was collected under realistic outdoor conditions, capturing diverse weather and lighting situations. The SuHiFiMask [1] dataset consists of three subsets: train, dev, and test, which consist of {159,063, 89,276, and 161,882} images, respectively. The images in each subset are assigned quality scores within specific ranges. Figure 2 provides a visual representation of some sample images from the dataset. The images in the train set exhibit high-quality facial features that are readily recognizable, while the dev set displays lower quality images, which are characterized by various types of noise, such as masks and occlusions. Conversely, the test set features the poorest quality images with the highest level of noise, including motion blur, lens flare, and low lighting conditions. Therefore, the FAS method we propose must possess a strong level of robustness to be effective in real-world surveillance settings where these types of noise are commonplace. To gain further insight into the dataset, we trained our baseline model and utilized it to generate a t-SNE ((t-Distributed Stochastic Neighbor Embedding) visualization of the features extracted from the train and dev sets. We observed that once the model had converged, it was able to accurately classify images from the train set. However, due to the lower quality of the images in the dev set, the model's performance experienced a slight decline.

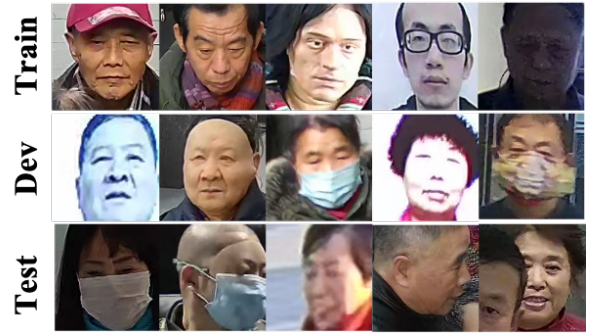

Figure 2. Samples of the SuHiFiMask dataset, which is partitioned based on face quality into three subsets: train, dev, and test.

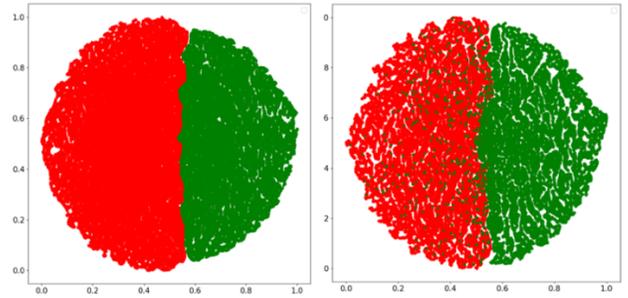

Figure 3. The t-SNE (t-Distributed Stochastic Neighbor Embedding) visualizations of the features extracted from both the train and dev sets.

#### 2) Evaluation Metrics

We utilize the most widely used metrics: Area Under the Curve (AUC) and Attack Classification Error Rate (ACER) to assess the efficacy of our proposed method. The AUC metric calculates the area under the Receiver Operating Characteristic (ROC) curve, which is a plot of the True Positive Rate (TPR) against the False Positive Rate (FPR) at various classification thresholds. The TPR is also known as sensitivity, which is the proportion of actual positives that are correctly classified as positive, while the FPR is the proportion of actual negatives that are incorrectly classified as positive. AUC is defined as follows:

$$AUC = \int_0^1 TPR(t)dFPR(t) \quad (3)$$

where $TPR(t)$ and $FPR(t)$ are the true positive rate and false positive rate, respectively. And $t$ is a given classification threshold.

The ACER is a metric used to assess the effectiveness of a FAS system in distinguishing between genuine and fake face presentations. It is calculated by averaging the attack presentation classification error rate (APCER) and the bona fide presentation classification error rate (BPCER), both measured at a specific decision threshold. The APCER indicates the rate at which presentation attacks are falsely identified as genuine face presentations, thereby indicating the level of security risk. On the other hand, the BPCER measures the rate at which genuine face presentations are mistakenly identified as presentation attacks, thus indicating the level of inconvenience to the user. By evaluating both types of errors, the ACER provides a comprehensive evaluation of the FAS system's performance. ACER is defined as follows:

$$ACER = (APCER + BPCER)/2 \qquad (4)$$

*3) Implementation Details*

Following [13], we use ViT-Large [5] as our backbone to extract feature embeddings. It was pretrained on ImageNet-1K dataset. All input images are resized to $224 \times 224 \times 3$ and normalized using the mean and standard deviation computed from SuHiFiMask dataset. Our training strategy involves using a batch size of 64 and stochastic gradient descent with momentum 0.9. The training process takes 80 epochs, with the first 3000 iterations using a warmup strategy. We set the initial learning rate to 0.01 and use Cosine decay to gradually decrease the learning rate. Besides, we employ low-quality augmentations such as motion blur, which is effective in scenarios. Additionally, we apply Test-Time Augmentation (TTA), which improves model accuracy and generalization. By using TTA, our models can handle variations in the test data, leading to better performance and generalization. To obtain the optimal performance, we conducted the final evaluation by utilizing the train and dev sets and computed the average scores on the test set with three iterations.

*B. Experimental Results*

TABLE I. COMPARING RESULTS ON TEST SET OF SUHIFIMASK DATASET

| Method | APCER | BPCER | ACER | AUC |
| --- | --- | --- | --- | --- |
| OPDAI | 9.18 | 5.13 | 7.16 | 97.38 |
| Horsego | **8.17** | **4.26** | **6.22** | 96.97 |
| Wida | 12.37 | 3.26 | 7.82 | 97.10 |
| Ours | 8.73 | 8.67 | 8.70 | **97.42** |

We conducted an extensive evaluation of our proposed DisE method and compared it with some techniques on the SuHiFiMask dataset. The results, which are presented in Table 1, clearly show that DisE achieves a highly competitive performance with an impressive AUC metric score of 97.42%. To further investigate the effectiveness of DisE, we conducted additional experiments and evaluated its impact on the original FAS model. As shown in Table 2, the results indicate that after implementing DisE, the model achieved a lower ACER and higher AUC scores, making it more robust and accurate. This is due to the fact that DisE allows the FAS model to learn from both clean and noisy (low-quality) input face images and adjust the learning strength of different quality samples. In summary, our proposed DisE method shows great potential for improving the robustness and accuracy of FAS systems, especially in challenging surveillance scenarios where low-quality face images are prevalent. By allowing the model to learn from both clean and noisy images, DisE enables the FAS model to better handle various types of noise and improve its overall performance. The results of our experiments demonstrate the effectiveness of DisE and its potential for real-world applications in the field of face anti-spoofing.

TABLE II. THE EFFECTIVENESS OF DISE AND TTA. EVALUATED ON TEST SET OF SUHIFIMASK DATASET

| Method | APCER | BPCER | ACER | AUC |
| --- | --- | --- | --- | --- |
| Baseline | 10.69 | 7.58 | 9.13 | 96.92 |
| w/ DisE | 8.94 | **8.61** | 8.77 | 97.04 |
| w/ DisE TTA | **8.73** | 8.67 | **8.70** | **97.42** |

V. Conclusions

In this work, we propose Distributional Estimation (DisE), a method that improves FAS stability and accuracy in long-distance surveillance by modeling data uncertainty during training. DisE introduces uncertainty in learned features, converting traditional point estimation to distributional estimation. Our evaluation on SuHiFiMask, a challenging FAS dataset, shows DisE's comparable performance on both ACER and AUC metrics. DisE has potential to improve the accuracy and robustness of FAS systems, particularly in challenging long-distance surveillance scenarios with low-quality face images.